\documentclass[runningheads]{llncs}
\usepackage{graphicx}
\usepackage{multirow}
\usepackage{booktabs}
\usepackage{adjustbox}
\usepackage{caption}
\usepackage{subcaption}
\usepackage{comment}
\usepackage{amsmath,amssymb} %
\usepackage{pgfplots}
\usepackage{tango}
\usepackage{cite}
\usepackage{wrapfig}
\usepackage{nicefrac}

\newcommand{\etal}{et al. }

\usepackage{amsmath}

\usepackage{xcolor}

\usepackage{xfrac}

\usepackage{array}
\newcolumntype{P}[1]{>{\centering\arraybackslash}p{#1}}
\newcolumntype{M}[1]{>{\centering\arraybackslash}m{#1}}
\newcolumntype{L}[1]{>{\raggedright\let\newline\\\arraybackslash}m{#1}}
\newcolumntype{C}[1]{>{\centering\let\newline\\\arraybackslash}m{#1}}
\newcolumntype{R}[1]{>{\raggedleft\let\newline\\\arraybackslash}m{#1}}

\usepackage{arydshln} %

\newcommand{\refsec}[1]{Section~\ref{sec:#1}}
\newcommand{\lblsec}[1]{\label{sec:#1}}
\newcommand{\reffig}[1]{Figure~\ref{fig:#1}}
\newcommand{\lblfig}[1]{\label{fig:#1}}
\newcommand{\reftbl}[1]{Table~\ref{tbl:#1}}
\newcommand{\lbltbl}[1]{\label{tbl:#1}}

\newcommand{\lbleq}[1]{\label{eq:#1}}

\makeatletter
\def\blfootnote{\xdef\@thefnmark{}\@footnotetext}
\makeatother

\usepackage[implicit=false,colorlinks=true,urlcolor=blue,frenchlinks=true]{hyperref}

\begin{document}
\pagestyle{headings}
\mainmatter
\def\ECCVSubNumber{5551}  %

\title{Domain Adaptation Through Task Distillation} %

\titlerunning{Domain Adaptation Through Task Distillation}
\author{Brady Zhou$^\ast$ \and %
Nimit Kalra$^\ast$ \and %
Philipp Kr\"ahenb\"uhl} %
\authorrunning{Brady Zhou, Nimit Kalra, and Philipp Kr\"ahenb\"uhl}
\institute{The University of Texas at Austin, Austin TX, USA\\
\email{\{bzhou, nimit, philkr\}@cs.utexas.edu}}
\maketitle

\blfootnote{$^\ast$ indicates equal contribution}

\begin{abstract}
Deep networks devour millions of precisely annotated images to build their complex and powerful representations.
Unfortunately, tasks like autonomous driving have virtually no real-world training data.
Repeatedly crashing a car into a tree is simply too expensive.
The commonly prescribed solution is simple: learn a representation in simulation and transfer it to the real world.
However, this transfer is challenging since simulated and real-world visual experiences vary dramatically.
Our core observation is that for certain tasks, such as image recognition, datasets are plentiful.
They exist in any interesting domain, simulated or real, and are easy to label and extend.
We use these recognition datasets to link up a source and target domain to transfer models between them in a task distillation framework.
Our method can successfully transfer navigation policies between drastically different simulators: ViZDoom, SuperTuxKart, and CARLA.
Furthermore, it shows promising results on standard domain adaptation benchmarks.
\end{abstract}

\section{Introduction}

Labeled data has been the main driving force behind the rise of deep learning~\cite{deng2009imagenet}.
Tasks with an abundance of labeled data flourished~\cite{deng2009imagenet,geiger2013vision,lin2014microsoft,cordts2016cityscapes,caesar2019nuscenes}, whereas tasks short on data saw only limited progress~\cite{saxena2008make3d,martin2001database}.
Over the past decade, deep networks have cut the error rate for image recognition by a factor of four~\cite{krizhevsky2012imagenet,ILSVRC15,he2016deep},
doubled object detection performance~\cite{felzenszwalb2009object,lin2014microsoft,he2017mask},
and enabled near-perfect pixel-wise segmentation of an image~\cite{chen2017rethinking,zhao2017pyramid}.
Yet, these same networks do not yet drive real-world autonomous vehicles, pilot a drone, or control a robot from the same diverse real-world visual inputs~\cite{caesar2019nuscenes}.
In simulation~\cite{ai2thor,dosovitskiy2017carla,savva2019habitat}, these tasks are not necessarily much harder to learn than recognition~\cite{chen2019learning} --- they simply have little to no labeled real-world data.
Unfortunately, models born and raised purely in simulation often fail to perform well in the real world~\cite{rusu2017sim,sadeghi2017cad2rl}.
This problem is not unique to simulated and real domains --- even models trained on a specific dataset often fail to generalize to other datasets~\cite{torralba2011unbiased}.

\begin{figure*}[t!]
    \centering
    \begin{subfigure}[t]{\linewidth}
    \centering
    \rotatebox[origin=bl]{90}{\ \  (a) Image domain}
      \includegraphics[width=0.31\textwidth,page=1]{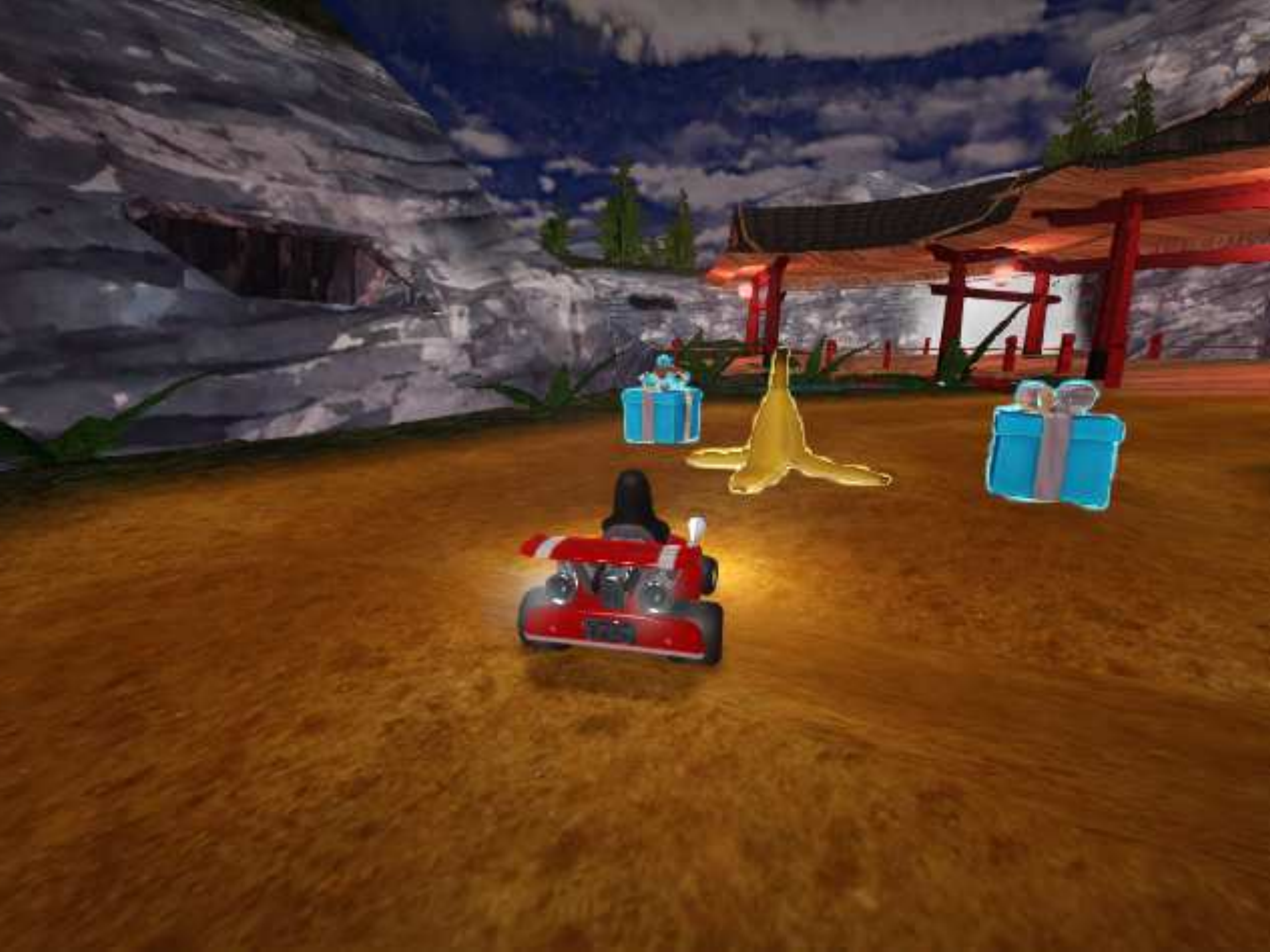}
      \includegraphics[width=0.31\textwidth,page=2]{figures/teaser/teaser.pdf}
      \includegraphics[width=0.31\textwidth,page=3]{figures/teaser/teaser.pdf}
    \end{subfigure}
    \begin{subfigure}[t]{\linewidth}
    \centering
    \rotatebox[origin=bl]{90}{\ \ \ (b) Label domain}
      \includegraphics[width=0.31\textwidth,page=5]{figures/teaser/teaser.pdf}
      \includegraphics[width=0.31\textwidth,page=6]{figures/teaser/teaser.pdf}
      \includegraphics[width=0.31\textwidth,page=8]{figures/teaser/teaser.pdf}
    \end{subfigure}
\caption{Raw visual inputs (a) may significantly vary across different domains, yet they often share common recognition labels (b). In this work, we use these recognition labels to transfer tasks between different domains.}
\lblfig{teaser}
\end{figure*}

Our core observation is that the gap between many datasets is much smaller in the output labels than the input images, as shown in \reffig{teaser}.
This is no accident.
Recognition tasks are carefully hand-designed to infer a compact, general, and abstract representation of the world~\cite{deng2009imagenet,everingham2010pascal,lin2014microsoft,OpenImages,shao2019objects365}.
Datasets often share largely overlapping label sets, and task definitions are compatible.
Most recognition tasks were designed as a first stepping stone to the rich world of visual reasoning tasks~\cite{zamir2018taskonomy,zhou2019does}.
Why not solve recognition in all domains, and then build downstream tasks on top of recognition models~\cite{muller2018driving,wong2019identifying,teichmann2018multinet,bansal2018chauffeurnet}?
Since the gap between label spaces is small, representations will generalize.
However, ``solving recognition'' turns out to be quite difficult~\cite{lin2014microsoft,caesar2019nuscenes,everingham2015pascal,ILSVRC15}.
Current recognition systems mislabel objects, detect an object where there is none, or worse, fail to recognize objects altogether.
If recognition is not solved in its entirety, errors will compound to downstream tasks, and a domain gap will persist between domains with good recognition systems and those with poor ones.

In this paper, we take a different approach.
We use the ground truth recognition labels directly to transfer downstream tasks from a source to target domain through task distillation.
First, we learn a proxy model that maps ground-truth recognition labels to outputs of the source model, through distillation~\cite{hinton2015distilling}.
This proxy model generalizes much better to the target domain, as it operates on a more compact and abstract input.
Next, we perform a second step of distillation to recover an image-based target model that imitates the proxy.
This target model, no longer uses any ground truth supervision and learns the task in an end-to-end manner.
We call this procedure \emph{task distillation} as it distills a source task to operate in a target domain with the help of an auxiliary recognition task.

This procedure may seem counterintuitive, but it has several advantages over other domain adaptation methods.
Firstly, recognition labels from different domains exhibit a smaller domain shift than their raw image counterparts.
Secondly, we do not need to solve recognition in either source or target domain --- we simply need a recognition dataset in each domain with a compatible label space.
Finally, task distillation results in an end-to-end model in the target domain, and does not suffer from compounding errors in deployment.

We investigate how our task transfer framework performs under two distinct domain adaptation applications: 1) driving policy transfer for visual navigation, and 2) simulation-to-reality transfer for semantic segmentation prediction.
We first show that our framework is able to transfer a lane-following driving policy from a simple racing game to a fully-fledged driving simulator.
Furthermore, our framework is even able to transfer an obstacle-avoidance policy from a maze-navigation video game to driving among other moving vehicles.
In the target domain, both of our transferred policies drive twice as far as the closest baselines.

Next, we apply our proposed framework to the standard domain adaptation task of transferring semantic segmentation models from simulated to real-world datasets.
Here, task distillation again significantly outperforms prior work.
Our framework is conceptually simple and easy to implement.
All code and data is publicly available at \href{https://github.com/bradyz/task-distillation}{\texttt{https://github.com/bradyz/task-distillation}}.

\section{Related Work}
Dataset bias is likely as old as machine learning itself~\cite{torralba2011unbiased} --- models trained on one dataset tend to generalize poorly out-of-the-box to related ones.
The rise of deep learning has ushered in a wave of massive datasets~\cite{deng2009imagenet,lin2014microsoft,cordts2016cityscapes,geiger2013vision,caesar2019nuscenes}.
Although these diverse datasets enable state-of-the-art models to better generalize to other datasets, a domain gap still exists.
One popular method to close this domain gap is to pre-train on a source domain and then fine-tune on a target domain~\cite{long2015fully,huh2016makes}.
In the same spirit, our work leverages a large amount of ground-truth labels to provide supervision for vision tasks in a different domain.
However, rather than rely on final task labels in the target domain, as they may be difficult or impossible to collect, we use generic recognition labels in both domains.

\paragraph{Domain adaptation} aims to bridge the source and target domain by adapting the weights of a model to increase its performance in a target domain.
Domain adaptation techniques include domain-specific normalization techniques~\cite{li2016revisiting},
statistical matching on input~\cite{zhu2017unpaired,murez2018image,vu2019advent}, output~\cite{hoffman2018cycada,tsai2018learning}, and intermediate activation~\cite{huang2018domain} distributions between source and target domains.
While most techniques rely only on the statistical distributions of the input data and the transferred model, recent works have introduced auxiliary labels and tasks to aid adaptation~\cite{sun2019unsupervised,vu2019dada}.
Ramirez \etal~\cite{ramirez2019learning} learn a common representation for an auxiliary task in both the source and target domain, then use this representation to link the two domains and aid transfer between them.
Our approach is significantly simpler and does not require training any models for the auxiliary task.

\paragraph{Simulation-to-reality transfer} has received a lot of attention in recent years, as simulators effortlessly produce massive amounts of labeled data~\cite{richter2017playing,krahenbuhl2018free}.
Transfer via modular pipelines is a promising simulation-to-reality method, wherein the observation space in both domains is mapped to a shared intermediate proxy task to ease generalization.
Doersch \etal~\cite{doersch2019sim2real} use motion to transfer 3D human pose labels.
M\"uller \etal~\cite{muller2018driving} use a semantic segmentation proxy task to transfer a driving policy, whereas Mousavian \etal~\cite{mousavian2019visual} use it as input to learned visual navigation policies.
Zhou \etal~\cite{zhou2019does} explore the impact of various high-level intermediate visual representations on learning to act, whereas Sax \etal~\cite{sax2020learning} explore how mid-level visual priors assist learning to navigate.

While modular systems all greatly outperform using unsupervised domain adaptation techniques, they still suffer from compounding errors.
If the intermediate auxiliary tasks is not solved perfectly, different error patterns between source and target domain will persist and lead to weaker transfer.
Our framework, on the other hand directly uses the error-free ground truth annotations for transfer, and thus does not suffer from compounding errors.

Another popular avenue for simulation-to-reality transfer is through the use of domain randomization.
In an effort to capture and generalize to the target domain distribution, these techniques randomize visual and dynamical properties of the simulation during training, similar to data augmentation in general deep learning.
Domain randomization requires no real-world labels and are especially popular in transferring robotic manipulation tasks~\cite{tobin2017domain,peng2018sim,akkaya2019solving,james2019sim}.
James \etal successfully use this technique to reduce the amount of real world training data for robot grasping by two orders of magnitude~\cite{james2019sim}.
While domain randomization works well in simple closed environments, it is not yet clear how to randomize more complex simulators to enable transfer to real-world visual domains.

\section{Method}

Our framework makes heavy use of model distillation~\cite{hinton2015distilling}.
Let $f_\theta(x)$  be a deep network with parameters $\theta$.
Let $g(x)$ be a second function, possibly another deep network.
Distillation trains $f_\theta$ to produce the same output as $g$ on a dataset $D$:
$$
\mathrm{minimize}_\theta  E_{x \sim D}\left[\ell(f_\theta(x), g(x))\right]
$$
Distillation learns to imitate more than just the ground truth labels. It gets to see and imitate all outputs from a target function $g$, and thus captures some of its inner workings and representation, also known as dark knowledge~\cite{hinton2015distilling}.
The original distillation work~\cite{hinton2015distilling} learns classification tasks using a smooth cross entropy loss $\ell$.
However, a large family of loss functions generally work.
For simplicity, we use an $L_1$ loss $\ell(f_\theta(x), g(x)) = |f_\theta(x) - g(x)|$ for both categorical and regression tasks.

Distillation is easily extended to a pair of models that use different inputs $x$ and $y$, as long as there exists a dataset with paired input modalities $(x,y)$.
$$
\mathrm{minimize}_\theta  E_{(x,y) \sim D}\left[\ell(f_\theta(x), g(y))\right]
$$

For notational simplicity, we call this process $f := \mathcal{D}_D(g)$ in later sections.
Distillation is the cornerstone of our domain adaptation algorithm, and the generalized process is also pervasive in policy optimization under the term behavior cloning or imitation learning~\cite{pomerleau1989alvinn}.
In previous works, it has been successfully used to replace a privileged driving policy with a pure sensorimotor policy~\cite{chen2019learning}.

\subsection{Task Distillation}

\begin{figure*}[t!]
\centering
\includegraphics[width=1.0\textwidth,page=1]{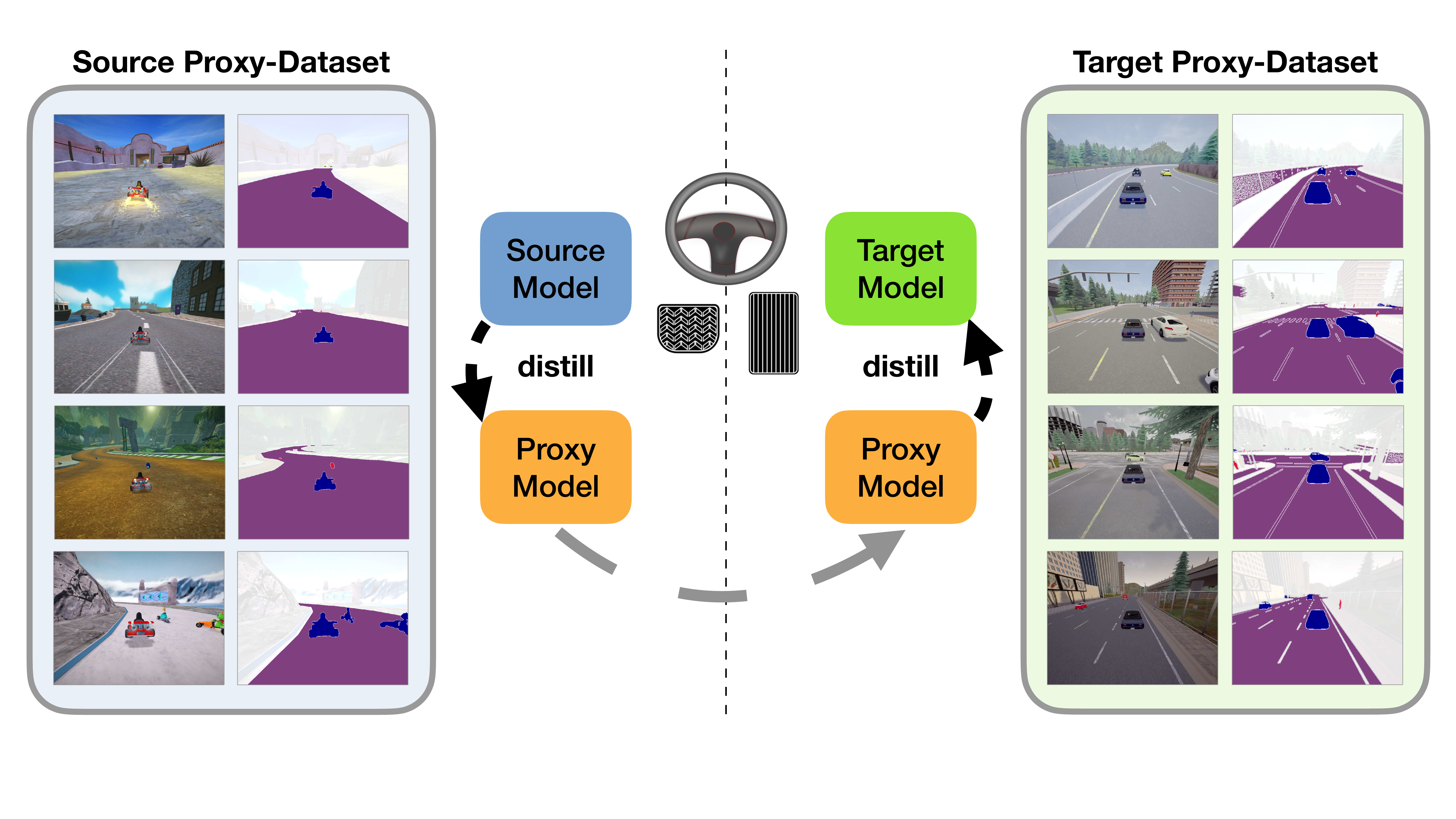}
\caption{Our method first distills a source model to a proxy model that uses labels as inputs. As proxy labels generalize to the target domain, a second stage of distillation is performed to produce a target model.}
\lblfig{method}
\end{figure*}

Let $\mathcal{S}$ be the source domain and $\mathcal{T}$ be the target domain.
We denote images from source and target domains as $I^S$ and $I^T$, respectively.
Let $L^S$ and $L^T$ be labels for a proxy task in both domains (e.g., image recognition).
Our goal is to transfer a model $f^S: I^S \to o$ producing outputs $o$ from the source domain to the target domain.
In our formulation, this output represents the desired prediction we aim to transfer to the target domain; hence, we assume $o$ is only available in the source domain.
We want to learn an adapted model $f^T: I^T \to o$ that performs the same task as $f^S$ in the target domain.

We propose a two-stage approach, as shown in \reffig{method}.
First, we learn a proxy model $f^P: L^S \to o$ using model distillation on a source dataset; i.e., $f^P := \mathcal{D}_\mathcal{S}(f^S)$.
Next, we distill the target model $f^T: I^T \to o$ from the proxy model on a target dataset, yielding $f^T := \mathcal{D}_\mathcal{T}(f^P)$.
If the label sets do not perfectly align between source and target domains, we bring them closer using simple hand designed transformations, such as remapping semantic labels, or different forms of data augmentation, as described in \refsec{results}.

\subsection{Comparison with Modular Approach}

Under which conditions does task distillation confer benefits over previous domain adaptation approaches?
Which scenarios will likely cause it to fail?
Both task distillation and the modular approach derive stronger generalization in the target domain through the use of an abstract, yet rich and informative, proxy task and proxy model.
However, whereas task distillation queries the proxy model at train-time with \textit{ground-truth} proxy labels, a modular pipeline queries this proxy model at deploy-time with \textit{predicted} proxy labels, thereby incurring the cost of an imperfect recognition system.

We denote the accuracy of the recognition system in the target domain as $a^l$, the proxy model accuracy in the source domain as $a^P$, and the \emph{similarity} between the two domains as $G^I = \sfrac{|I^S \cap I^T|}{|I^T|}$ in image space and $G^L = \sfrac{|L^S \cap L^T|}{|L^T|}$ in label space. Suppose a failure at any stage causes the entire system to fail. Then, the final accuracy is
\begin{equation}
a_\mathrm{modular}^T = a^P a^l G^L\lbleq{modular_acc};
\end{equation}
that is, the system succeeds only if proxy, recognition, and label transfer succeed.

Similarly, let $a^d$ be the accuracy of the second distillation stage in our proposed framework. Then, task distillation succeeds at a rate of
\begin{equation}
a_\mathrm{distill}^T = a^P G^L a^d\lbleq{distill_acc};
\end{equation}
that is, when the labels transfer and the second distillation succeeds.

Finally, direct transfer ignores the issue of domain shift altogether, and simply evaluates the source model $f^S$ in the target image domain. If $f^S$ has accuracy $a^S$, this approach succeeds at a rate of
\begin{equation}
a_\mathrm{direct}^T = a^S G^I\lbleq{direct_acc}.
\end{equation}

These accuracy estimations can be difficult to compare without references or experiments.
However, note that experimentally distillation commonly does not lose any accuracy --- Hinton \etal~\cite{hinton2015distilling} observe an increase in accuracy through distillation, while Chen \etal~\cite{chen2019learning} show equivalent accuracies.
It is thus safe to assume that $a^S \approx a^P$ in most cases.

With these estimates, we can reason about the relationships between domain adaptation approaches and develop some intuition.
Firstly, if the domain gap in the image domain is not significantly larger than the domain gap in the label domain ($G^I \approx G^L$), then direct transfer is likely to work quite well.
Moreover, task distillation and the modular approach differ by a single term --- the recognition accuracy $a^l$ versus the distillation accuracy $a^d$.
If recognition is easier to learn in the target domain, a modular approach likely yields a higher accuracy.
However, if the target task is easier to learn through distillation, our task distillation approach likely works better.

We find that, for many applications, modular pipelines must infer a proxy recognition task that is often far richer than needed for the end task.
Target tasks are empirically easier to learn than recognition, as they can often be inferred from a subset of recognition.
Although recognition is easier to supervise, with ground-truth labels in abundance, solving recognition pixel-perfect is very difficult.
In contrast, the target task of driving from pixels, for example, only relies on inferring a subset of recognition.
Intuitively, imperfect recognition 100m down the road, or on the opposite lane of a separated freeway, does not impact downstream driving performance.

We note that our task distillation framework and the modular approach both rely on $a^P$ and $G^L$ --- their success depends on the performance of the proxy model in solving the final task and the ability of the proxy task to generalize between domains.
Hence, the choice of proxy task is a careful trade-off between expressiveness and abstractness.

Although this analysis provides some intuition, our assumptions do ignore several practical aspects of modular approaches and distillation.
Firstly, not all mistakes are equally bad --- models are generally able to recover from partial failures and degrade slowly.
Moreover, while the second stage of our task distillation framework may introduce some mislabeled training data if the label spaces are too dissimilar, deep learning algorithms (namely stochastic gradient descent with data augmentation) gracefully generalize and can withstand some erroneous supervision.
Hence, these accuracy estimations may be overly pessimistic.

\begin{figure*}[t]
	\setlength{\tabcolsep}{5pt}
	\centering
	\setlength\dashlinedash{6pt}
	\setlength\dashlinegap{3pt}
	\setlength\arrayrulewidth{0.8pt}
	\def\arraystretch{1.5}
	\begin{tabular}{C{0.17\textwidth}C{0.18\textwidth}:C{0.18\textwidth}C{0.17\textwidth}C{0.18\textwidth}}
		\centering Source Input & Ground-Truth & Ground-Truth & \centering Predicted & Target Input \\
		
		\includegraphics[height=0.06\textheight]{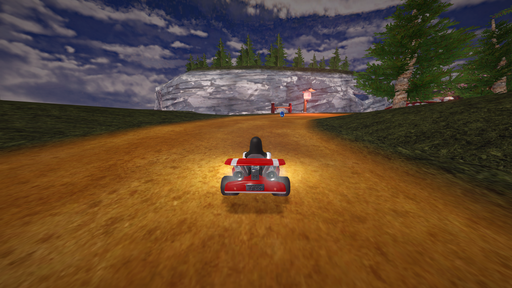} &
		\includegraphics[height=0.06\textheight]{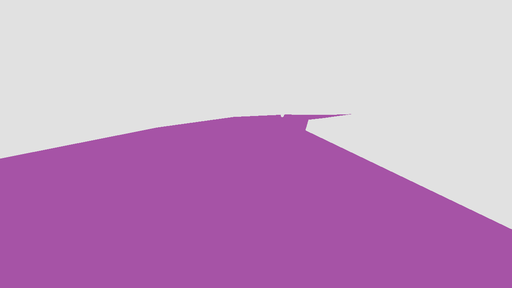} &
		\includegraphics[height=0.06\textheight]{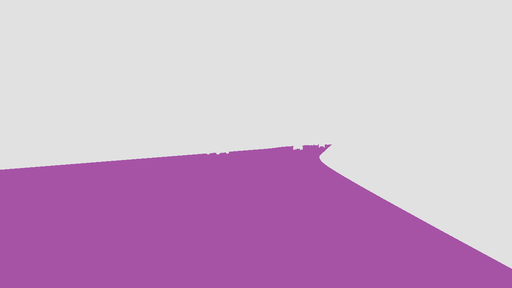} &
		\includegraphics[height=0.06\textheight]{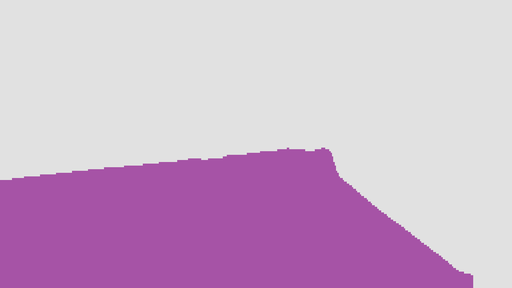} &
		\includegraphics[height=0.06\textheight]{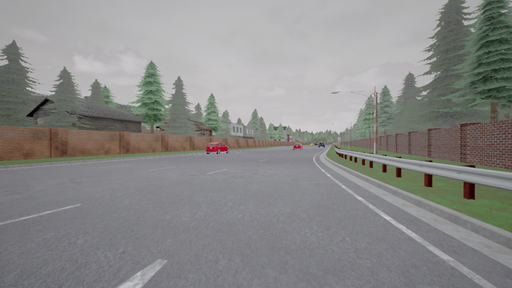} \\
		
		\includegraphics[height=0.06\textheight]{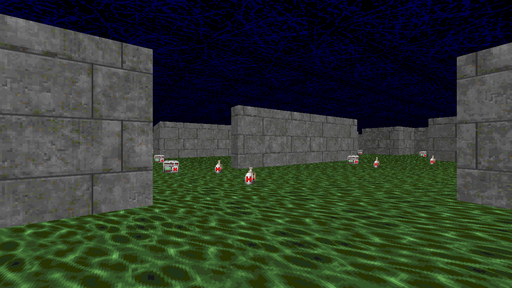} &
		\includegraphics[height=0.06\textheight]{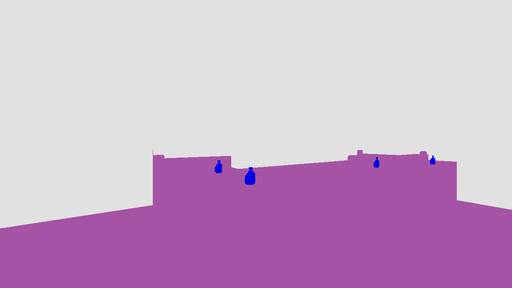} &
		\includegraphics[height=0.06\textheight]{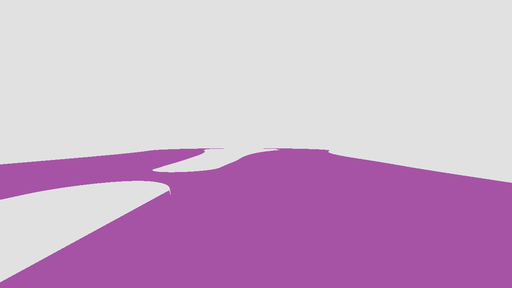} &
		\includegraphics[height=0.06\textheight]{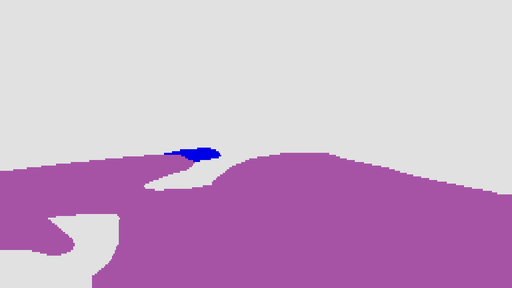} &
		\includegraphics[height=0.06\textheight]{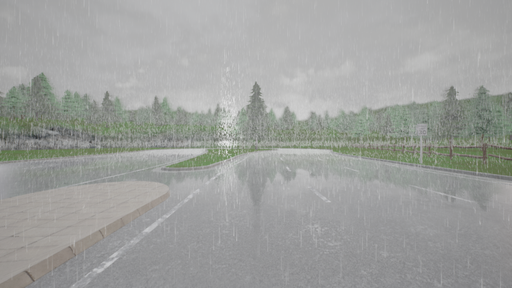} \\
		
		\includegraphics[height=0.06\textheight]{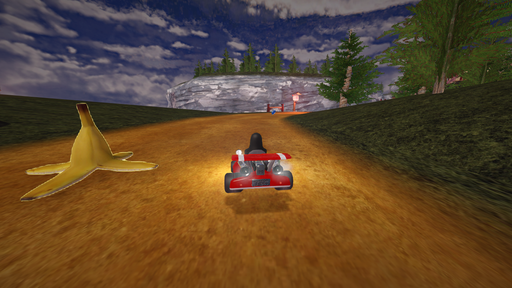} &
		\includegraphics[height=0.06\textheight]{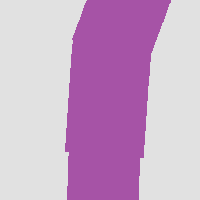} &
		\includegraphics[height=0.06\textheight]{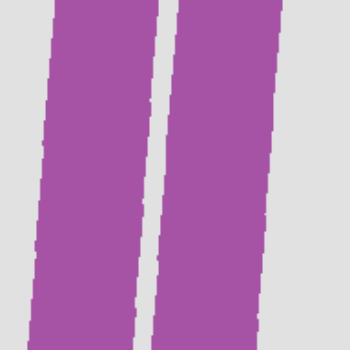} &
		\includegraphics[height=0.06\textheight]{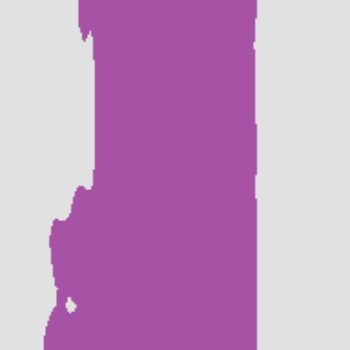} &
		\includegraphics[height=0.06\textheight]{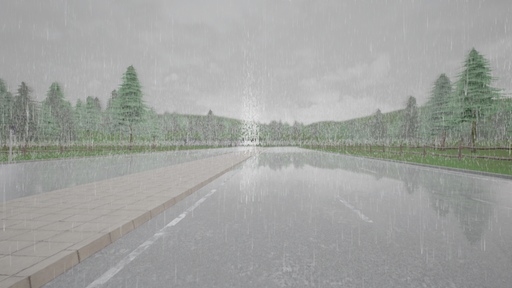} \\
		
		\includegraphics[height=0.06\textheight]{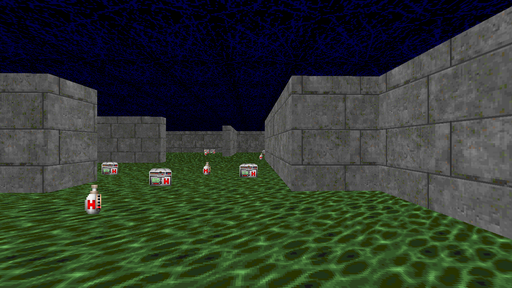} &
		\includegraphics[height=0.06\textheight]{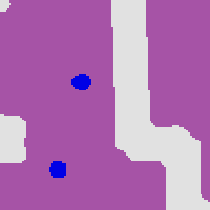} &
		\includegraphics[height=0.06\textheight]{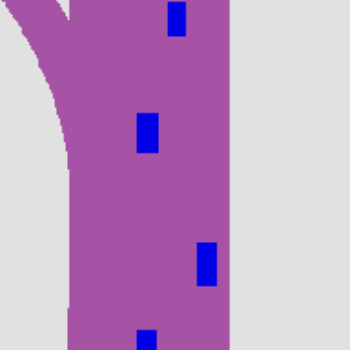} &
		\includegraphics[height=0.06\textheight]{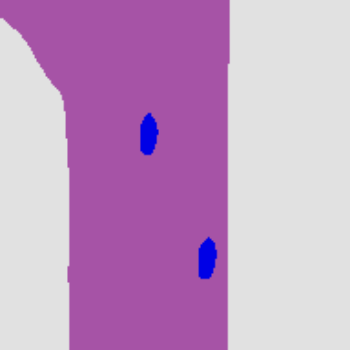} &
		\includegraphics[height=0.06\textheight]{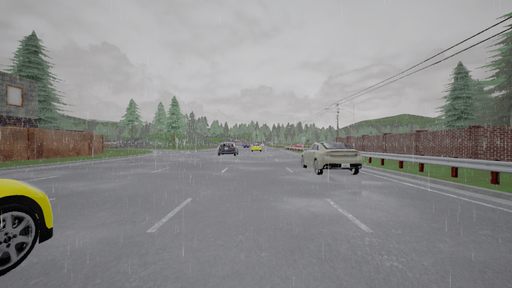} \\
		
		\multicolumn{2}{c}{(a) Source Domain} &
		\multicolumn{3}{c}{(b) Target Domain}
	\end{tabular}
	\caption{We compare visual domains by their raw monocular images and corresponding semantic representations. While the domains vary significantly in their raw images, they are quite similar in their semantic modalities. However, note that the predicted modalities used by a modular pipeline are not perfect. For example, in the bottom-most row, the map-view prediction fails to capture the yellow car in view directly left of the agent. When supplied to the downstream driving policy, this vision failure can result in unintended behavior.}
	\label{fig:modalities}
\end{figure*}

\section{Experiments and Results}
\lblsec{results}

We demonstrate our task distillation framework in two domain adaptation scenarios: 1) policy transfer of a navigation policy, and 2) general domain adaptation for semantic segmentation prediction.

\begin{figure*}[t]
	\setlength{\tabcolsep}{0pt}
	\centering
	\begin{tabular}{M{0.07\textwidth}M{0.2325\textwidth}M{0.2325\textwidth}M{0.2325\textwidth}M{0.2325\textwidth}}
		\tabcolsep0em
		& Direct & CyCADA & Modular & Ours \\
		\rotatebox[origin=bc]{90}{\parbox{1.5cm}{\linespread{0.1}\selectfont \centering SuperTux map-view}} &
		\includegraphics[width=0.2325\textwidth]{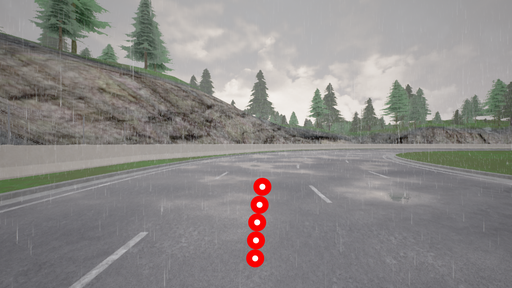} &
		\includegraphics[width=0.2325\textwidth]{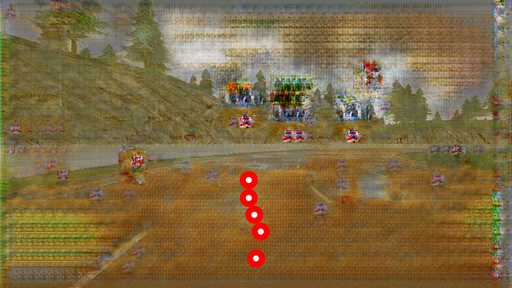} &
		\includegraphics[width=0.2325\textwidth]{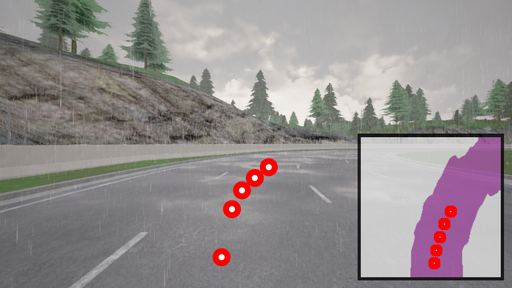} &
		\includegraphics[width=0.2325\textwidth]{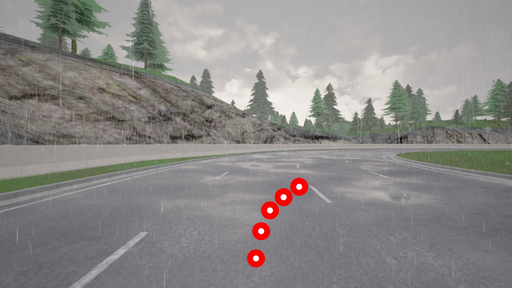} \\
		
		\rotatebox[origin=bl]{90}{\parbox{1.8cm}{\linespread{0.1}\selectfont \centering SuperTux cam-view}} &
		\includegraphics[width=0.2325\textwidth]{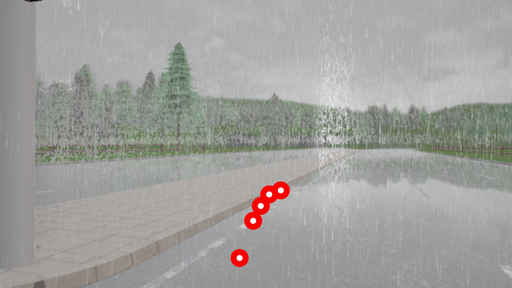} &
		\includegraphics[width=0.2325\textwidth]{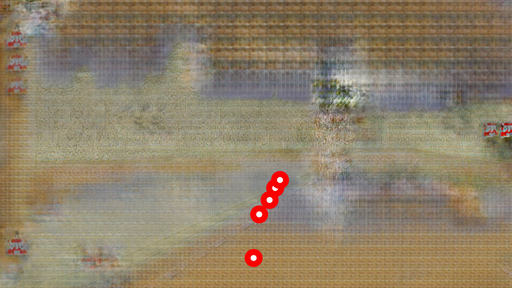} &
		\includegraphics[width=0.2325\textwidth]{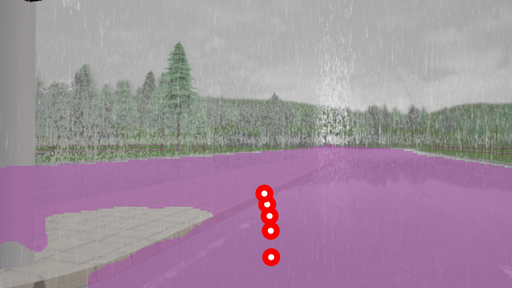} &
		\includegraphics[width=0.2325\textwidth]{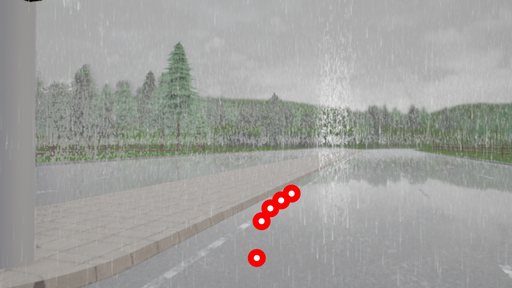} \\
		
		\rotatebox[origin=bl]{90}{\parbox{1.5cm}{\linespread{0.1}\selectfont \centering ViZDoom map-view}} &
		\includegraphics[width=0.2325\textwidth]{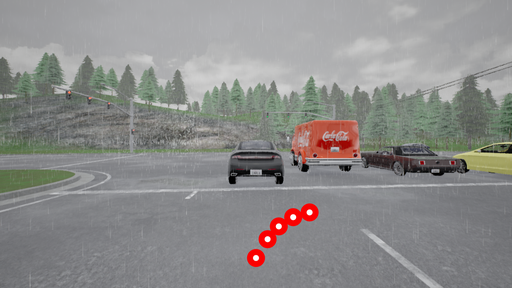} &
		\includegraphics[width=0.2325\textwidth]{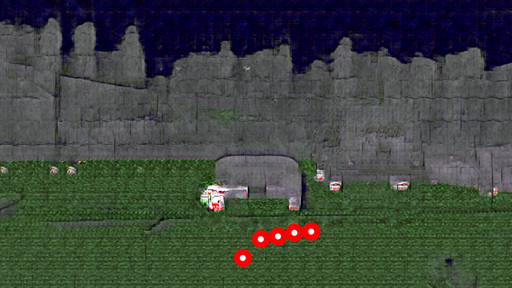} &
		\includegraphics[width=0.2325\textwidth]{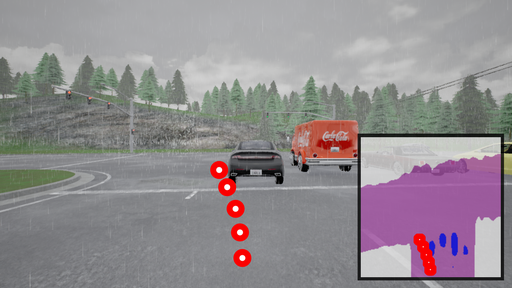} &
		\includegraphics[width=0.2325\textwidth]{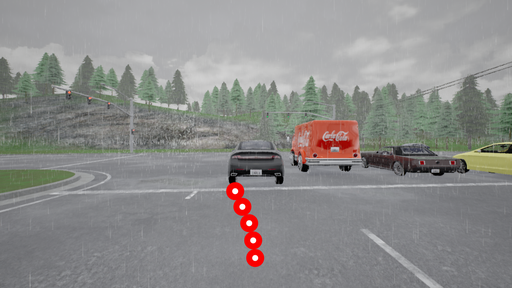} \\
		
		\rotatebox[origin=bl]{90}{\parbox{1.8cm}{\linespread{0.1}\selectfont \centering ViZDoom cam-view}} &
		\includegraphics[width=0.2325\textwidth]{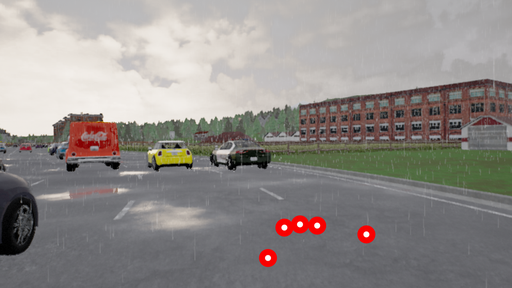} &
		\includegraphics[width=0.2325\textwidth]{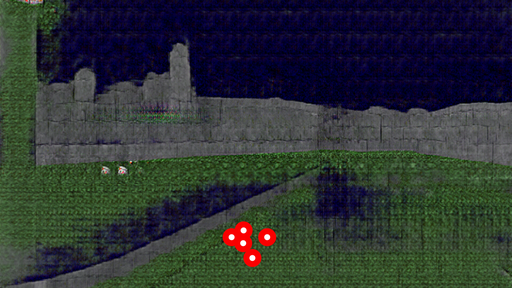} &
		\includegraphics[width=0.2325\textwidth]{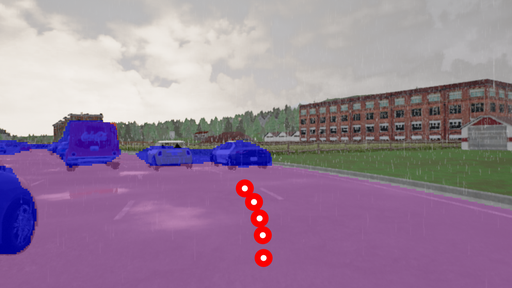} &
		\includegraphics[width=0.2325\textwidth]{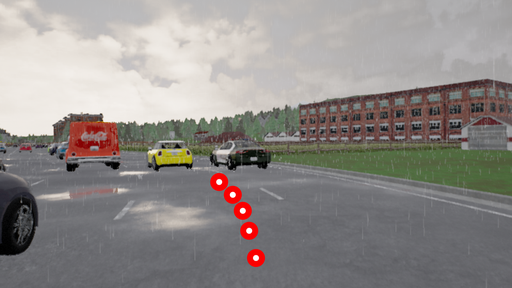} \\
	\end{tabular}
	\caption{We qualitatively examine how four different driving policies transfer to CARLA. Each policy is evaluated at the same state over four transfer methods, with predicted waypoints shown in red. Inferred modality is displayed for CyCADA and Modular. As shown, an inaccurate modality is used by a modular driving policy when transferring from SuperTuxKart via camera-view semantic segmentation. The median is misclassified as drivable road and the predicted waypoints direct the agent off of the road. (Best viewed on screen.)}
	\label{fig:predicted}
\end{figure*}

\subsection{Policy Transfer}
Evaluating real-world navigational policies in a controlled and reproducible fashion can be tricky, requiring a physical vehicle and a reusable testbed environment.
We sidestep these issues and instead transfer a navigation policy between three simulators of drastically different input fidelity, physical accuracy, and complexity: 1) SuperTuxKart, a simplistic open-source racing game, 2) ViZDoom~\cite{kempka2016vizdoom}, a professionally-developed maze-based shooting game, and 3) CARLA~\cite{dosovitskiy2017carla}, a photorealistic driving simulator.
As shown in \reffig{modalities}, the visual domain shift between these simulators is significant.
We train our policies in the relatively low-fidelity ViZDoom and SuperTuxKart video games and transfer these agents to drive in the realistic CARLA driving simulator.

We build our policies on the network architecture of Chen \etal~\cite{chen2019learning}, a ResNet-18 backbone that regresses to a trajectory plan of waypoints, which provide a domain-agnostic abstraction for control~\cite{muller2018driving}.
During deployment, a low-level PID controller converts waypoints to steering, throttle, and brake controls in CARLA.

We chose semantic segmentation in either camera-view~\cite{muller2018driving} or map-view~\cite{wang2019monocular,bansal2018chauffeurnet,chen2019learning} (also known as bird's-eye view in the literature) as the proxy recognition representation.
Semantic segmentation is a particularly useful proxy task since it is extremely prevalent in most datasets and domains.
Moreover, it allows us to easily enforce relationships between domains simply by mapping source classes to target classes.
In doing so, we can easily frame the desired behavior of the target CARLA vehicle in terms of the source agent we wish to transfer.

We build our policies on the network architecture of Chen \etal~\cite{chen2019learning}, wherein each policy outputs a trajectory plan using waypoints.
This representation provides a domain-agnostic abstraction for control~\cite{muller2018driving}.
During deployment in the target domain, waypoints are fed into a low-level PID controller to obtain steering, throttle, and brake controls in CARLA.

\begin{table*}[t]
	\center
	\begin{tabular}{l c r c c c c c c}
		\toprule
		Method & Proxy task & \multicolumn{3}{c}{Distance traveled (m)} & \multicolumn{4}{c}{Completion rate} \\
		& & \multicolumn{1}{c}{avg.} & min & max & 100m & 250m & 500m & 1000m \\
		\midrule
		Direct & --- & $22.4$ \scriptsize$\pm$3.2~ & 16.6 & 26.0 & 0.00 & 0.00 & 0.00 & 0.00 \\
		CyCADA~\cite{hoffman2018cycada} & --- & $24.0$ \scriptsize$\pm$1.3~ & 22.4 & 26.0 & 0.00 & 0.00 & 0.00 & 0.00 \\
		CyCADA$^\dagger$ & --- & $26.7$ \scriptsize$\pm$2.0~ & 23.6 & 28.7 & 0.02 & 0.00 & 0.00 & 0.00 \\
		\midrule
		Modular$^{\mathsection}$ & cam-view & $89.9$ \scriptsize$\pm$9.8~ & 81.4 & 108.6 & 0.24 & 0.08 & 0.00 & 0.00 \\
		Modular~\cite{muller2018driving} & cam-view & $110.4$ \scriptsize$\pm$17.1 & 95.7 & 138.2 & 0.38 & 0.06 & 0.02 & 0.01 \\
		Ours & cam-view & $164.6$ \scriptsize$\pm$14.9 & 147.5 & 191.6 & 0.59 & 0.18 & 0.03 & 0.00 \\

		\midrule
		Modular$^{\mathsection}$ & map-view & ~$49.9$ \scriptsize$\pm$3.8~ & 42.7 & 52.82 & 0.11 & 0.03 & 0.00 & 0.00 \\
		Modular & map-view & $135.3$ \scriptsize$\pm$8.0~ & 126.0 & 147.3 & 0.44 & 0.12 & 0.04 & 0.00 \\
		Ours & map-view & $\textbf{260.5}$ \scriptsize$\pm$15.2 & \textbf{ 244.5 } & \textbf{ 281.3 } & \textbf{ 0.66 } & \textbf{ 0.26 } & \textbf{ 0.20 } & \textbf{ 0.03 } \\
		\bottomrule
	\end{tabular}
	\caption{Adapting a SuperTuxKart racing agent to perform lane-following in CARLA. For each method, we evaluate 25 episodes using five fixed PID controller parameters. $\dagger$ denotes transferring raw low-level steering and throttle control to CARLA, as opposed to waypoints. ${\mathsection}$ denotes training the driving policy using proxy task predictions in the source domain, as opposed to ground-truth labels.}
	\lbltbl{supertux2carla}
\end{table*}

\paragraph{Evaluation.} For each transferred policy, we evaluate how far the agent travels until a driving infraction (i.e., collision) in an unseen test town in CARLA.
We force all agents to travel at $20$ \sfrac{km}{h}, as traveling speed greatly impacts the agent's performance.
For waypoint-based agents, we hand-tune five PID controllers on a reference planner in a training town, and use these controllers to obtain low-level driving commands.
We then evaluate all agents for 25 episodes under each controller, selecting the weather configuration and spawn points at random.
Our experimental setup follows the official evaluation protocol of CARLA~\cite{dosovitskiy2017carla}, except we do not have a fixed goal, and thus do not use high-level commands.
Instead, agents can chose to follow any route through the test town.

\paragraph{Baselines.} We evaluate our method against direct transfer, image-to-image translation via CyCADA~\cite{hoffman2018cycada}, and modular transfer~\cite{muller2018driving}.
Direct transfer ignores the issue of domain shift --- we simply evaluate the source model in the target domain.
Different baselines rely on varying proxy supervision during training, but the final model for each method maps raw input images to waypoints.

\begin{table*}[t]
	\center
	\begin{tabular}{l c r c c c c c c}
		\toprule
		Method & Proxy task & \multicolumn{3}{c}{Distance traveled (m)} & \multicolumn{4}{c}{Completion rate} \\
		& & \multicolumn{1}{c}{avg.} & min & max & 100m & 250m & 500m & 1000m \\
		\midrule
		Direct & --- & $17.4$ \scriptsize$\pm$2.4~ & 13.6 & 21.0 & 0.02 & 0.00 & 0.00 & 0.00 \\
		CyCADA & --- & $24.4$ \scriptsize$\pm$5.1~ & 20.7 & 33.9 & 0.02 & 0.00 & 0.00 & 0.00 \\
		\midrule
		Modular$^{\mathsection}$  & cam-view & $148.6$ \scriptsize$\pm$40.3 & 92.6 & 211.2 & 0.42 & 0.20 & 0.01 & 0.00 \\
		Modular & cam-view & $140.4$ \scriptsize$\pm$18.6 & 120.2 & 173.7 & 0.51 & 0.16 & 0.02 & 0.00 \\
		Ours & cam-view  & $166.9$ \scriptsize$\pm$33.6 & 125.2 & 223.8 & 0.62 & 0.17 & 0.06 & 0.00 \\
		\midrule
		Modular$^{\mathsection}$ & map-view & $89.5$ \scriptsize$\pm$7.8~ & 78.3 & 100.9 & 0.31 & 0.06 & 0.00 & 0.00 \\
		Modular & map-view & $145.3$ \scriptsize$\pm$15.5 & 125.2 & 170.9 & 0.55 & 0.18 & 0.02 & 0.00 \\
		Ours & map-view & $\textbf{277.3}$ \scriptsize$\pm$56.6 & \textbf{ 204.2 } & \textbf{ 353.3 } & \textbf{ 0.63 } & \textbf{ 0.35 } & \textbf{ 0.20 } & \textbf{ 0.05 } \\
		\bottomrule
	\end{tabular}
	\caption{Results from transferring a ViZDoom maze-navigation agent to perform lane-following and vehicle-avoidance in CARLA.}
	\lbltbl{vizdoom2carla}
\end{table*}

\paragraph{SuperTuxKart $\to$ CARLA.}
SuperTuxKart is a simple racing game, wherein players are expected to race on winding tracks --- which can be quite challenging to traverse --- by controlling the steering angle and throttle.
Compared to the two-way traffic and major intersections found in CARLA environments, each SuperTuxKart map has a single non-overlapping one-way track.
We train our source policy in SuperTuxKart using proximal policy optimization (PPO)~\cite{schulman2017proximal} to maximize the distance traveled down the track.
Our goal is to transfer this source policy to drive in CARLA.
Following the setup of M\"uller \etal~\cite{muller2018driving}, we disable other traffic participants in CARLA, and focus solely navigating the road.
We report our results in \reftbl{supertux2carla}.

Ignoring the issue of visual domain shift and simply deploying our source policy in CARLA results in predictably poor results --- the agent drives $100$m without infraction only 2\% of the time.
Attempting to translate the target visual inputs to emulate the style of the source domain (as done in unsupervised domain adaptation approaches such as CyCADA) yields a poor and inconsistent inferred modality, as shown in \reffig{predicted}.
Using this faulty input representation results in an agent with similarly low driving performance.
As highlighted in \reffig{modalities}, these two visual domains are simply too far from each other.

By leveraging the compact semantic segmentation representation, the modular pipeline approach and our task distillation method both achieve an order of magnitude improvement in driving performance.
For both sets of experiments, we align the ``track'' class in SuperTuxKart with the ``road'' class in CARLA to preserve the original source task's semantic relationship with the navigable track region (i.e., to stay on the track).

We observe that the map-view proxy task works better for both methods.
As Wang \etal~\cite{wang2019monocular} show, this representation is more informative than the camera-view, but it is also harder to infer due to perspective distortion.
Task distillation enables us to transfer a policy that is able to navigate twice as far as a modular approach using the same map-view proxy task.
Moreover, the map-view results in a $66.2$\% increase in distance traveled over semantic camera-view under task distillation; however, it provides almost no improvement in a modular pipeline.
Finally, unlike with a modular pipeline, our agent can exploit a stronger signal with being distracted by distortions, as they do not exist in the ground truth maps.
See \reffig{modalities} for visual examples.

\setlength\intextsep{0pt}
\begin{wrapfigure}{R}{0.5\textwidth}
	\includegraphics[width=0.975\linewidth]{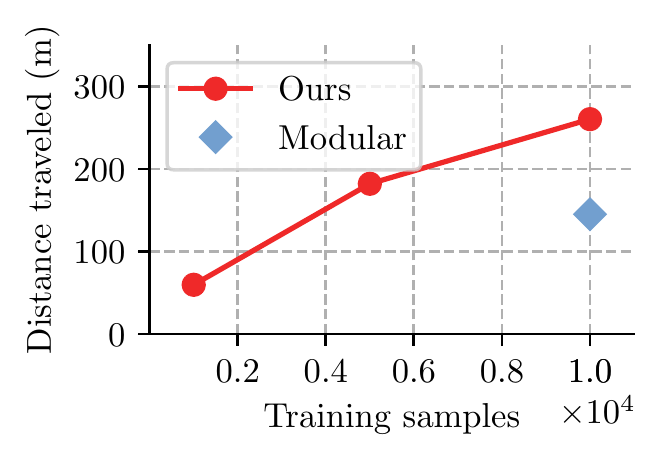}
	\setlength{\abovecaptionskip}{-5pt}
	\caption{Performance at different amounts of target-domain training data.}
	\lblfig{datasetsize}
\end{wrapfigure}
Finally, we evaluate the performance of our transferred agent as the target-domain proxy-task dataset size changes (see \reffig{datasetsize}).
By default, we use ten thousand labeled target images to both train the second stage of task distillation and to train the target recognition model in our modular pipelines.
We find that even when training on only \textit{half} of the target-domain samples, task distillation outperforms the modular approach.

\paragraph{ViZDoom $\to$ CARLA.}
In ViZDoom, we train an agent using direct future prediction (DFP) \cite{dosovitskiy2016learning} to navigate and explore complex maze environments while searching for health-kits and avoiding poison.
Unlike the winding SuperTuxKart tracks, ViZDoom maze corridors resemble the road intersections of an urban driving environment.
We aim to transfer this policy to a crowded street scenario in CARLA, wherein the agent must avoid other traffic participants while also navigating the road.
To this end, we semantically align the two domains by mapping the ``poison'' class in ViZDoom with the ``car'' class in CARLA, in addition to aligning the ``road'' and ``floor'' classes.
Hence, we frame the agent's tendency to avoid poison in the context of our target domain; i.e., the transferred policy avoids cars while staying on the road.
We report our results in \reftbl{vizdoom2carla}.

Direct transfer and CyCADA fail to yield a strong target policy and crash within a few meters.
Despite facing a more difficult CARLA evaluation with traffic participants, both the modular approach and task distillation outperform their counterpart experiments in SuperTuxKart.
We attribute these improvements to the inherent similarity between ViZDoom corridors and urban roads.
Task distillation significantly exceeds the performance of a modular pipeline, especially when using the expressive, but difficult to infer, map-view proxy task.

\begin{figure*}[t]
	\setlength{\tabcolsep}{1pt}
	\centering
	\begin{tabular}{cccc}
		\tabcolsep0em
		Input & Depth & Ours & Ground-Truth \\
		\includegraphics[width=0.23\textwidth]{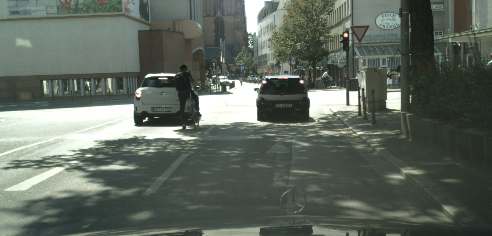} &
		\includegraphics[width=0.23\textwidth]{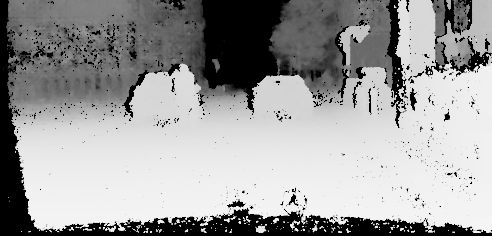} &
		\includegraphics[width=0.23\textwidth]{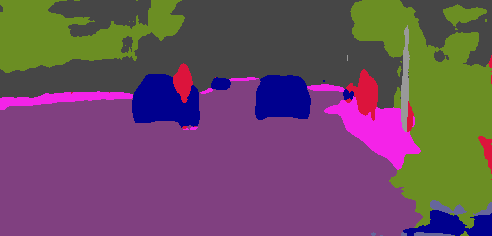} &
		\includegraphics[width=0.23\textwidth]{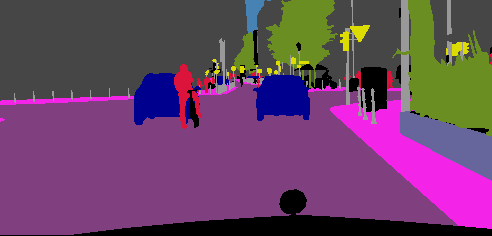} \\
		
		\includegraphics[width=0.23\textwidth]{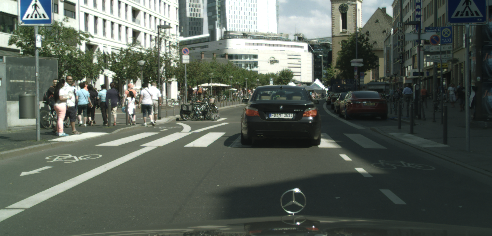} &
		\includegraphics[width=0.23\textwidth]{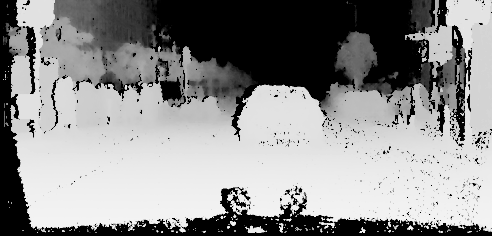} &
		\includegraphics[width=0.23\textwidth]{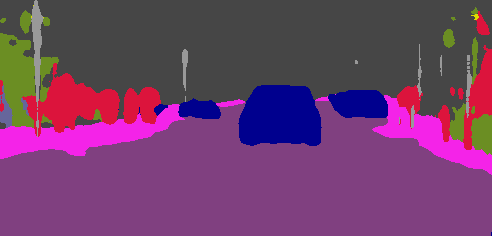} &
		\includegraphics[width=0.23\textwidth]{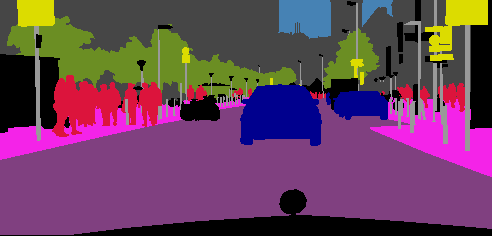} \\
		
		\includegraphics[width=0.23\textwidth]{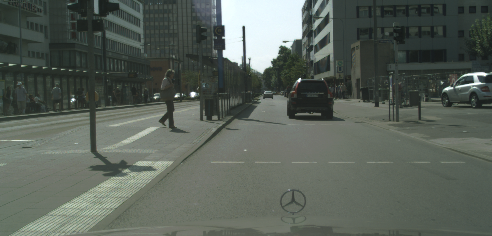} &
		\includegraphics[width=0.23\textwidth]{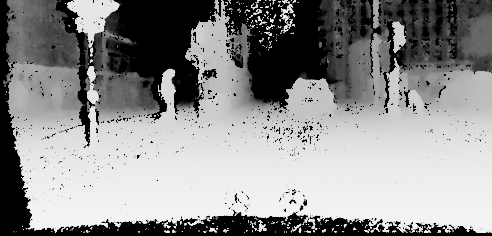} &
		\includegraphics[width=0.23\textwidth]{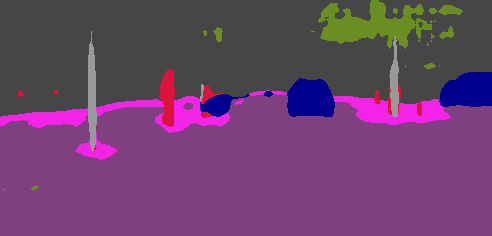} &
		\includegraphics[width=0.23\textwidth]{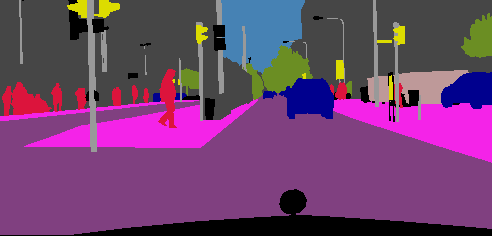} \\
		
		\includegraphics[width=0.23\textwidth]{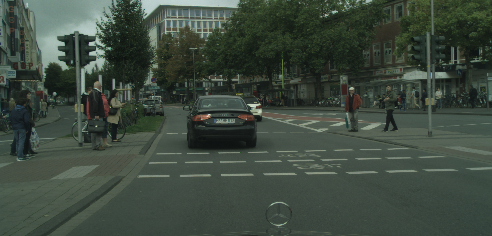}&
		\includegraphics[width=0.23\textwidth]{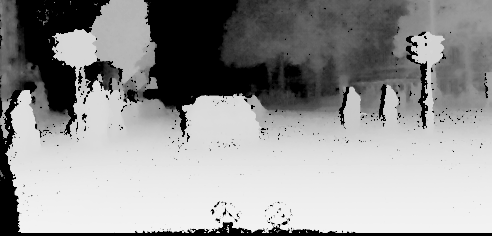}  &
		\includegraphics[width=0.23\textwidth]{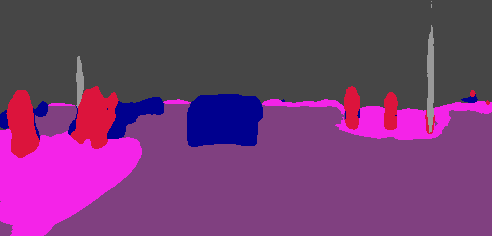} &
		\includegraphics[width=0.23\textwidth]{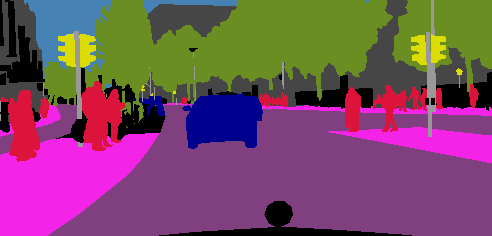} \\
		
		\includegraphics[width=0.23\textwidth]{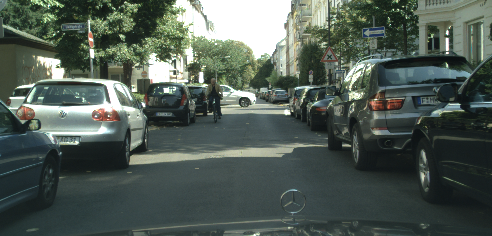} &
		\includegraphics[width=0.23\textwidth]{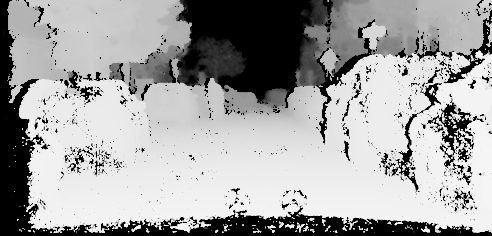} &
		\includegraphics[width=0.23\textwidth]{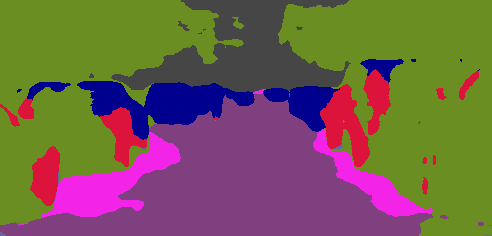} &
		\includegraphics[width=0.23\textwidth]{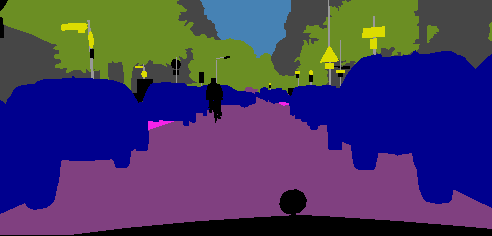} \\
		
		\includegraphics[width=0.23\textwidth]{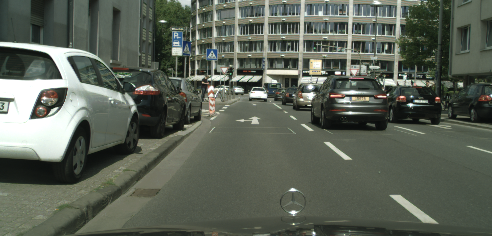}&
		\includegraphics[width=0.23\textwidth]{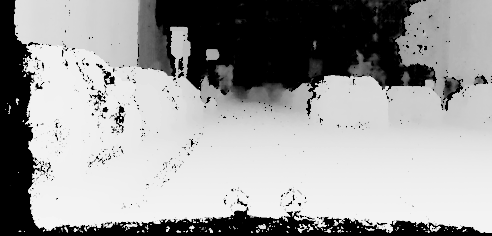} &
		\includegraphics[width=0.23\textwidth]{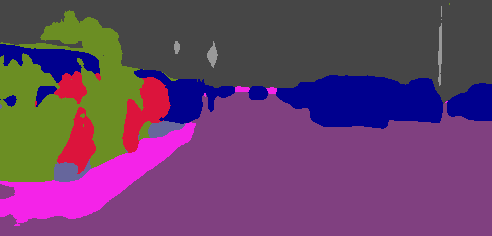} &
		\includegraphics[width=0.23\textwidth]{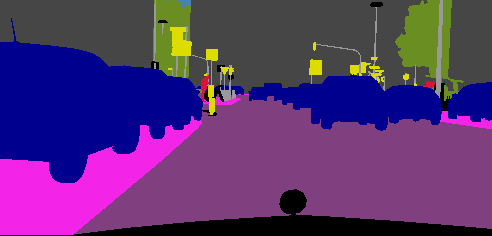}  \\
	\end{tabular}
	\caption{Transferring the semantic segmentation task from the simulated SYNTHIA-SF dataset to the real-world Cityscapes dataset using task distillation with a depth estimation proxy task. We display representative samples near the reported mIoU in the top four rows. When depth estimation from stereo vision is too noisy, performance drops substantially, as shown in the bottom two rows. This highlights the importance of proxy task performance. Our final adapted model predicts semantic segmentation using only raw monocular images --- we simply display the proxy depth labels for illustration.}
	\lblfig{sem_qualitative}
\end{figure*}

Our policy transfer experiments concern a very specific domain adaptation application.
However, task distillation is not limited to transferring sequential decision-making policies between domains.
We can apply our framework to transfer general computer vision tasks from simulated to real-world datasets.

\subsection{General Domain Adaptation}

We transfer the semantic segmentation task between a variety of datasets: 1) SYNTHIA~\cite{ros2016synthia,HernandezBMVC17}, a collection of synthetic scenes in a virtual city, 2) Cityscapes~\cite{cordts2016cityscapes}, an assortment of video sequences recorded on real-world urban streets, and 3) a hand-collected set of frames rendered in the CARLA driving simulator.
All three datasets consist of raw monocular images with semantic and depth annotations.
We train a semantic segmentation prediction model on the SYNTHIA-SF dataset and adapt the model on both the Cityscapes and CARLA datasets.

In our task distillation setup, we use the low-level visual signal of depth estimation as a proxy task.
Our experimental setup mimics that of Ramirez \etal~\cite{ramirez2019learning}, who use depth as an auxiliary supervisory signal to link domains in their AT/DT framework.
However, we utilize a different semantic segmentation network architecture, building upon the simple DeepLabv3 model~\cite{chen2017rethinking}.

\paragraph{Evaluation.}
As these datasets may have incompatible semantic classes, we merge these classes as per Ramirez \etal~\cite{ramirez2019learning}.
We use standard evaluation metrics for semantic segmentation, and report the intersection-over-union per class, mean IoU, and global pixel-wise accuracy.
Although we rely on proxy labels during training, just as in our policy transfer experiments, our final adapted models predict semantic segmentation from raw images in the target domain.

\begin{table*}[t]
	\center
	\setlength{\tabcolsep}{2.5pt}
	\scalebox{0.65}{
		\begin{tabular}{lccl|ccccccccccc|cc}
			\toprule
			& Source & Target & Method & \rotatebox{90}{Road} & \rotatebox{90}{Sidewalk} & \rotatebox{90}{Walls} & \rotatebox{90}{Fence} & \rotatebox{90}{Person} & \rotatebox{90}{Poles} & \rotatebox{90}{Vegetation} & \rotatebox{90}{Vehicles} & \rotatebox{90}{Tr. Signs} & \rotatebox{90}{Building}  & \rotatebox{90}{Sky} & \textbf{mIoU} & \textbf{Acc} \\
			\midrule
			\multirow{4}{*}{\textbf{(a)}}
			& SYNTHIA-SF & CARLA & Direct & 9.1 & 30.9 & 0.1 & 0.0 & 13.7 & 14.9 & 57.3 & 15.5 & 4.5 & 22.8 & 64.6 & 21.2 & 48.0 \\
			& SYNTHIA-SF & CARLA & AT/DT$^{\mathparagraph}$ & 63.9 & 54.9 & 15.2 & 0.0 & 13.6 & 12.8 & 52.7 & 27.3 & 4.9 & 50.2 & 79.7 & 34.1 & 73.4 \\
			& SYNTHIA-SF & CARLA & AT/DT~\cite{ramirez2019learning} & 73.6 & 62.6 & \textbf{26.9} & 0.0 & 17.8 & 37.3 & 35.3 & 52.9 & 17.8 & \textbf{63.0} & 87.5 & 43.1 & 80.0 \\
			& SYNTHIA-SF & CARLA & Ours & \textbf{95.4} & \textbf{90.2} & 10.4 & 0.0 & \textbf{26.6} & \textbf{52.1} & \textbf{66.1} & \textbf{60.5} & \textbf{53.1} & 57.3 & \textbf{95.5} & \textbf{55.2} & \textbf{87.0} \\
			
			\midrule
			\multirow{3}{*}{\textbf{(b)}}
			& SYNTHIA-SF & Cityscapes & Direct & 0.7 & 1.6 & 0.0 & 0.0 & 13.4 & 6.0 & \textbf{57.4} & 21.6 & \textbf{2.0} & \underline{42.7} & 19.7 & 15.0 & 41.3 \\
			& SYNTHIA-SF & Cityscapes & AT/DT$^{\mathparagraph}$ & 6.9 & 0.7 & 0.0 & 0.0 & 2.5 & 9.1 & 3.2 & 8.9 & 0.8 & 25.9 & 26.9 & 7.7 & 28.5 \\
			& SYNTHIA-SF & Cityscapes & AT/DT & 85.8 & 29.4 & 1.2 & 0.0 & 3.7 & 14.6 & 1.9 & 8.9 & 0.4 & \textbf{42.8} & \textbf{67.1} & 23.2 & \underline{64.0} \\
			& SYNTHIA-SF & Cityscapes & Ours & \textbf{86.8} & \textbf{46.3} & \textbf{8.1} & 0.0 & \textbf{34.9} & \textbf{19.2} & 12.9 & \textbf{52.4} & 0.3 & 40.9 & 6.4 & \textbf{28.0} & \textbf{64.3} \\
			\midrule
			\midrule
			& SYNTHIA-SF & Cityscapes & Proxy$^\ast$ & 84.9 & 37.5 & 8.2 & 0.0 & 26.6 & 23.0 & 16.8 & 43.1 & 19.3 & 43.3 & 19.7 & 29.3 & 64.4 \\
			\bottomrule
		\end{tabular}
	}
	\caption{Transferring a semantic segmentation prediction model between simulated and real-world datasets. ${\mathparagraph}$ denotes using the AT/DT architecture for direct transfer. ~$^\ast$ denotes evaluating our proxy model on the target dataset.}
	\lbltbl{depth2sem}
\end{table*}

\paragraph{SYNTHIA-SF $\to$ CARLA.}
As shown in \reftbl{depth2sem}, the AT/DT direct baseline (denoted by $\mathparagraph$) outperforms our direct baseline.
As their baseline uses no auxiliary tasks for aiding transfer, this seems to suggest that our model architecture is inferior.
However, task distillation is able to easily bridge this difference without any architectural changes.
Using the same additional depth supervision, our simple DeepLabv3 model yields significant improvement in mIoU over the complex AT/DT architecture.
In particular, we observe increase in the IoU of several pervasive semantic classes: ``roads'', ``sidewalk'', ``vegetation'', and ``traffic sign.''

\paragraph{SYNTHIA-SF $\to$ Cityscapes.}
Since the Cityscapes dataset contains noisy depth estimations from stereo images, whereas the SYNTHIA-SF dataset provides depth labels directly from the rendering engine, there is a non-negligible domain gap in label space.
Moreover, stereo matching failures produce gaps in the Cityscapes depth estimation.
Hence, we must carefully align the proxy labels to ensure strong transfer.
To this end, we apply various data augmentations to make the SYNTHIA-SF depth maps more similar to the noisy Cityscapes labels.
During training, we simulate this noise by randomly sampling Cityscapes masks and applying stereo-matching failure regions to the perfect depth maps in the source datasets.
Furthermore, to encourage invariance to slight scale changes, we add per-pixel multiplicative noise to each image.

Despite this proxy label mismatch between the two datasets, our adapted model achieves an absolute improvement of $4.8$ mIoU over Ramirez \etal~\cite{ramirez2019learning}, yielding improved accuracy in the ``sidewalk'', ``person'', and ``vehicle'' semantic classes, as shown in \reffig{sem_qualitative}.
However, we struggle with the ``sky'' class, likely due to noisy depth labels.
This result is quite exciting, as we do not depend on any specialized real-world sensors.
Using simple stereo images, we can transfer a semantic segmentation model trained in simulation for use in the real world.

Our adapted model performs well on scenes with high-quality depth estimation and struggles otherwise.
As our method thrives when the domain gap between label spaces is small, and hence better depth estimates would likely further improve final performance.
Moreover, our final model performs similarly to the proxy model in the target domain, thereby indicating that 1) the final distillation is successful, and 2) our final model's performance is constrained by the inability of our proxy model to generalize to the noisy proxy labels.

\paragraph{Limitations.} Task distillation, like the modular approach, relies heavily on the proxy task being transferable.
For example, adapting a model from SYNTHIA-RAND, in which the camera poses vary drastically, to Cityscapes results in an mIoU of $11.5$.
Transfer via the depth estimation proxy task fails due to the inconsistent appearance induced by changes in camera pose.

\section{Conclusion}
In this work, we propose a simple and effective framework for transferring knowledge from a source to target domain.
Our method doesn't require any end-task labels in the target domain.
Instead, we choose a proxy task with ample annotations in both simulation and the real world, e.g., depth or semantics, to tie both domains together.
Task distillation outperforms competing alternatives for simulation-to-reality navigation policy transfer and domain adaptation for semantic segmentation.

\paragraph{Acknowledgments.}
This work has been supported in part by the National Science Foundation under grant IIS-1845485.

\par\vfill\par

\clearpage
\bibliographystyle{splncs04}
\bibliography{egbib}

\begin{thebibliography}{10}
\providecommand{\url}[1]{\texttt{#1}}
\providecommand{\urlprefix}{URL }
\providecommand{\doi}[1]{https://doi.org/#1}

\bibitem{akkaya2019solving}
Akkaya, I., Andrychowicz, M., Chociej, M., Litwin, M., McGrew, B., Petron, A.,
  Paino, A., Plappert, M., Powell, G., Ribas, R., et~al.: Solving {Rubik's}
  cube with a robot hand. arXiv preprint arXiv:1910.07113  (2019)

\bibitem{bansal2018chauffeurnet}
Bansal, M., Krizhevsky, A., Ogale, A.: {ChauffeurNet}: Learning to drive by
  imitating the best and synthesizing the worst. In: RSS (2019)

\bibitem{caesar2019nuscenes}
Caesar, H., Bankiti, V., Lang, A.H., Vora, S., Liong, V.E., Xu, Q., Krishnan,
  A., Pan, Y., Baldan, G., Beijbom, O.: {nuScenes}: A multimodal dataset for
  autonomous driving. arXiv preprint arXiv:1903.11027  (2019)

\bibitem{chen2019learning}
Chen, D., Zhou, B., Koltun, V., Kr{\"a}henb{\"u}hl, P.: Learning by cheating.
  In: CoRL (2019)

\bibitem{chen2017rethinking}
Chen, L.C., Papandreou, G., Schroff, F., Adam, H.: Rethinking atrous
  convolution for semantic image segmentation. arXiv preprint arXiv:1706.05587
  (2017)

\bibitem{cordts2016cityscapes}
Cordts, M., Omran, M., Ramos, S., Rehfeld, T., Enzweiler, M., Benenson, R.,
  Franke, U., Roth, S., Schiele, B.: The {Cityscapes} dataset for semantic
  urban scene understanding. In: CVPR (2016)

\bibitem{deng2009imagenet}
Deng, J., Dong, W., Socher, R., Li, L.J., Li, K., Fei-Fei, L.: {ImageNet}: A
  large-scale hierarchical image database. In: CVPR (2009)

\bibitem{doersch2019sim2real}
Doersch, C., Zisserman, A.: Sim2real transfer learning for 3d human pose
  estimation: motion to the rescue. In: NeurIPS (2019)

\bibitem{dosovitskiy2016learning}
Dosovitskiy, A., Koltun, V.: Learning to act by predicting the future. In: ICLR
  (2017)

\bibitem{dosovitskiy2017carla}
Dosovitskiy, A., Ros, G., Codevilla, F., Lopez, A., Koltun, V.: {CARLA}: An
  open urban driving simulator. In: CoRL (2017)

\bibitem{everingham2015pascal}
Everingham, M., Eslami, S.A., Van~Gool, L., Williams, C.K., Winn, J.,
  Zisserman, A.: The {PASCAL} visual object classes challenge: A retrospective.
  In: IJCV (2015)

\bibitem{everingham2010pascal}
Everingham, M., Van~Gool, L., Williams, C.K., Winn, J., Zisserman, A.: The
  {PASCAL} visual object classes ({VOC}) challenge. In: IJCV (2010)

\bibitem{felzenszwalb2009object}
Felzenszwalb, P.F., Girshick, R.B., McAllester, D., Ramanan, D.: Object
  detection with discriminatively trained part-based models. In: TPAMI (2009)

\bibitem{geiger2013vision}
Geiger, A., Lenz, P., Stiller, C., Urtasun, R.: Vision meets robotics: The
  {KITTI} dataset. In: IJRR (2013)

\bibitem{he2017mask}
He, K., Gkioxari, G., Doll{\'a}r, P., Girshick, R.: Mask {R-CNN}. In: ICCV
  (2017)

\bibitem{he2016deep}
He, K., Zhang, X., Ren, S., Sun, J.: Deep residual learning for image
  recognition. In: CVPR (2016)

\bibitem{HernandezBMVC17}
Hernandez-Juarez, D., Schneider, L., Espinosa, A., Vazquez, D., Lopez, A.M.,
  Franke, U., Pollefeys, M., Moure, J.C.: Slanted stixels: Representing san
  francisco’s steepest streets. In: BMVC (2017)

\bibitem{hinton2015distilling}
Hinton, G., Vinyals, O., Dean, J.: Distilling the knowledge in a neural
  network. arXiv preprint arXiv:1503.02531  (2015)

\bibitem{hoffman2018cycada}
Hoffman, J., Tzeng, E., Park, T., Zhu, J.Y., Isola, P., Saenko, K., Efros, A.,
  Darrell, T.: {CyCADA}: Cycle-consistent adversarial domain adaptation. In:
  ICML (2018)

\bibitem{huang2018domain}
Huang, H., Huang, Q., Kr{\"a}henb{\"u}hl, P.: Domain transfer through deep
  activation matching. In: ECCV (2018)

\bibitem{huh2016makes}
Huh, M., Agrawal, P., Efros, A.A.: What makes {ImageNet} good for transfer
  learning? arXiv preprint arXiv:1608.08614  (2016)

\bibitem{james2019sim}
James, S., Wohlhart, P., Kalakrishnan, M., Kalashnikov, D., Irpan, A., Ibarz,
  J., Levine, S., Hadsell, R., Bousmalis, K.: Sim-to-real via sim-to-sim:
  Data-efficient robotic grasping via randomized-to-canonical adaptation
  networks. In: CVPR (2019)

\bibitem{kempka2016vizdoom}
Kempka, M., Wydmuch, M., Runc, G., Toczek, J., Ja{\'s}kowski, W.: {ViZDoom}: A
  {Doom}-based {AI} research platform for visual reinforcement learning. In:
  CIG (2016)

\bibitem{ai2thor}
Kolve, E., Mottaghi, R., Han, W., VanderBilt, E., Weihs, L., Herrasti, A.,
  Gordon, D., Zhu, Y., Gupta, A., Farhadi, A.: {AI2-THOR}: An interactive {3D}
  environment for visual {AI}. arXiv preprint arXiv:1712.05474  (2017)

\bibitem{krahenbuhl2018free}
Kr{\"a}henb{\"u}hl, P.: Free supervision from video games. In: CVPR (2018)

\bibitem{krizhevsky2012imagenet}
Krizhevsky, A., Sutskever, I., Hinton, G.E.: {ImageNet} classification with
  deep convolutional neural networks. In: NeurIPS (2012)

\bibitem{OpenImages}
Kuznetsova, A., Rom, H., Alldrin, N., Uijlings, J., Krasin, I., Pont-Tuset, J.,
  Kamali, S., Popov, S., Malloci, M., Duerig, T., Ferrari, V.: {The Open Images
  Dataset V4}: Unified image classification, object detection, and visual
  relationship detection at scale. arXiv preprint arXiv:1811.00982  (2018)

\bibitem{li2016revisiting}
Li, Y., Wang, N., Shi, J., Liu, J., Hou, X.: Revisiting batch normalization for
  practical domain adaptation. In: ICLR (2017)

\bibitem{lin2014microsoft}
Lin, T.Y., Maire, M., Belongie, S., Hays, J., Perona, P., Ramanan, D.,
  Doll{\'a}r, P., Zitnick, C.L.: Microsoft {COCO}: Common objects in context.
  In: ECCV (2014)

\bibitem{long2015fully}
Long, J., Shelhamer, E., Darrell, T.: Fully convolutional networks for semantic
  segmentation. In: CVPR (2015)

\bibitem{martin2001database}
Martin, D., Fowlkes, C., Tal, D., Malik, J.: A database of human segmented
  natural images and its application to evaluating segmentation algorithms and
  measuring ecological statistics. In: ICCV (2001)

\bibitem{mousavian2019visual}
Mousavian, A., Toshev, A., Fi{\v{s}}er, M., Ko{\v{s}}eck{\'a}, J., Wahid, A.,
  Davidson, J.: Visual representations for semantic target driven navigation.
  In: ICRA (2019)

\bibitem{muller2018driving}
M{\"u}ller, M., Dosovitskiy, A., Ghanem, B., Koltun, V.: Driving policy
  transfer via modularity and abstraction. In: CoRL (2018)

\bibitem{murez2018image}
Murez, Z., Kolouri, S., Kriegman, D., Ramamoorthi, R., Kim, K.: Image to image
  translation for domain adaptation. In: CVPR (2018)

\bibitem{peng2018sim}
Peng, X.B., Andrychowicz, M., Zaremba, W., Abbeel, P.: Sim-to-real transfer of
  robotic control with dynamics randomization. In: ICRA (2018)

\bibitem{pomerleau1989alvinn}
Pomerleau, D.A.: {ALVINN}: An autonomous land vehicle in a neural network. In:
  NeurIPS (1989)

\bibitem{ramirez2019learning}
Ramirez, P.Z., Tonioni, A., Salti, S., Stefano, L.D.: Learning across tasks and
  domains. In: ICCV (2019)

\bibitem{richter2017playing}
Richter, S.R., Hayder, Z., Koltun, V.: Playing for benchmarks. In: ICCV (2017)

\bibitem{ros2016synthia}
Ros, G., Sellart, L., Materzynska, J., Vazquez, D., Lopez, A.M.: The {SYNTHIA}
  dataset: A large collection of synthetic images for semantic segmentation of
  urban scenes. In: CVPR (2016)

\bibitem{ILSVRC15}
Russakovsky, O., Deng, J., Su, H., Krause, J., Satheesh, S., Ma, S., Huang, Z.,
  Karpathy, A., Khosla, A., Bernstein, M., Berg, A.C., Fei-Fei, L.: {ImageNet
  Large Scale Visual Recognition Challenge}. In: IJCV (2015)

\bibitem{rusu2017sim}
Rusu, A.A., Vecerik, M., Roth{\"o}rl, T., Heess, N., Pascanu, R., Hadsell, R.:
  Sim-to-real robot learning from pixels with progressive nets. In: CoRL (2017)

\bibitem{sadeghi2017cad2rl}
Sadeghi, F., Levine, S.: {CAD2RL}: Real single-image flight without a single
  real image. In: RSS (2017)

\bibitem{savva2019habitat}
Savva, M., Kadian, A., Maksymets, O., Zhao, Y., Wijmans, E., Jain, B., Straub,
  J., Liu, J., Koltun, V., Malik, J., et~al.: Habitat: A platform for embodied
  {AI} research. In: ICCV (2019)

\bibitem{sax2020learning}
Sax, A., Zhang, J.O., Emi, B., Zamir, A., Savarese, S., Guibas, L., Malik, J.:
  Learning to navigate using mid-level visual priors. In: CoRL (2020)

\bibitem{saxena2008make3d}
Saxena, A., Sun, M., Ng, A.Y.: {Make3D}: Learning {3D} scene structure from a
  single still image. In: TPAMI (2008)

\bibitem{schulman2017proximal}
Schulman, J., Wolski, F., Dhariwal, P., Radford, A., Klimov, O.: Proximal
  policy optimization algorithms. arXiv preprint arXiv:1707.06347  (2017)

\bibitem{shao2019objects365}
Shao, S., Li, Z., Zhang, T., Peng, C., Yu, G., Zhang, X., Li, J., Sun, J.:
  Objects365: A large-scale, high-quality dataset for object detection. In:
  ICCV (2019)

\bibitem{sun2019unsupervised}
Sun, Y., Tzeng, E., Darrell, T., Efros, A.A.: Unsupervised domain adaptation
  through self-supervision. ICLR  (2019)

\bibitem{teichmann2018multinet}
Teichmann, M., Weber, M., Zoellner, M., Cipolla, R., Urtasun, R.: {MultiNet}:
  Real-time joint semantic reasoning for autonomous driving. In: Intelligent
  Vehicles (2018)

\bibitem{tobin2017domain}
Tobin, J., Fong, R., Ray, A., Schneider, J., Zaremba, W., Abbeel, P.: Domain
  randomization for transferring deep neural networks from simulation to the
  real world. In: IROS (2017)

\bibitem{torralba2011unbiased}
Torralba, A., Efros, A.A.: Unbiased look at dataset bias. In: CVPR (2011)

\bibitem{tsai2018learning}
Tsai, Y.H., Hung, W.C., Schulter, S., Sohn, K., Yang, M.H., Chandraker, M.:
  Learning to adapt structured output space for semantic segmentation. In: CVPR
  (2018)

\bibitem{vu2019advent}
Vu, T.H., Jain, H., Bucher, M., Cord, M., P{\'e}rez, P.: {ADVENT}: Adversarial
  entropy minimization for domain adaptation in semantic segmentation. In: CVPR
  (2019)

\bibitem{vu2019dada}
Vu, T.H., Jain, H., Bucher, M., Cord, M., P{\'e}rez, P.: {DADA}: Depth-aware
  domain adaptation in semantic segmentation. In: ICCV (2019)

\bibitem{wang2019monocular}
Wang, D., Devin, C., Cai, Q.Z., Kr{\"a}henb{\"u}hl, P., Darrell, T.: Monocular
  plan view networks for autonomous driving. In: ICRA (2019)

\bibitem{wong2019identifying}
Wong, K., Wang, S., Ren, M., Liang, M., Urtasun, R.: Identifying unknown
  instances for autonomous driving. In: CoRL (2019)

\bibitem{zamir2018taskonomy}
Zamir, A.R., Sax, A., Shen, W., Guibas, L.J., Malik, J., Savarese, S.:
  Taskonomy: Disentangling task transfer learning. In: CVPR (2018)

\bibitem{zhao2017pyramid}
Zhao, H., Shi, J., Qi, X., Wang, X., Jia, J.: Pyramid scene parsing network.
  In: CVPR (2017)

\bibitem{zhou2019does}
Zhou, B., Kr{\"a}henb{\"u}hl, P., Koltun, V.: Does computer vision matter for
  action? In: Science Robotics (2019)

\bibitem{zhu2017unpaired}
Zhu, J.Y., Park, T., Isola, P., Efros, A.A.: Unpaired image-to-image
  translation using cycle-consistent adversarial networks. In: ICCV (2017)

\end{thebibliography}
\end{document}